\documentclass[preprint,12pt]{elsarticle}

\usepackage[utf8]{inputenc}
\usepackage[T1]{fontenc}
\usepackage{amsmath,amssymb,amsfonts}
\usepackage{booktabs}
\usepackage{multirow}
\usepackage{graphicx}
\usepackage{xcolor}
\usepackage{hyperref}
\usepackage{cleveref}
\usepackage{subcaption}
\usepackage{array}
\usepackage{tabularx}
\usepackage{adjustbox}
\usepackage{enumitem}
\usepackage{xspace}
\usepackage{siunitx}

\graphicspath{{figures/}}

\newcommand{\pp}{\,pp\xspace}

\newcommand{\etal}{et~al.\xspace}

\newcommand{\modelqwen}{Qwen3.5-0.8B\xspace}
\newcommand{\modelfalcon}{Falcon-H1-0.5B\xspace}

\hypersetup{
  colorlinks=true,
  linkcolor=blue!70!black,
  citecolor=blue!70!black,
  urlcolor=blue!70!black
}

\begin{document}

\begin{frontmatter}

\title{Where Should LoRA Go? Component-Type Placement in Hybrid Language Models}

\author[vrain]{H\'ector Borobia\corref{cor1}}
\ead{hecboar@doctor.upv.es}

\author[econ]{Elies Segu\'i-Mas}
\author[org]{Guillermina Tormo-Carb\'o}

\cortext[cor1]{Corresponding author}
\affiliation[vrain]{organization={VRAIN -- Valencian Research Institute for Artificial Intelligence, Universitat Polit\`ecnica de Val\`encia},
  city={Valencia},
  country={Spain}}
\affiliation[econ]{organization={Department of Economics and Social Sciences, Universitat Polit\`ecnica de Val\`encia},
  city={Valencia},
  country={Spain}}
\affiliation[org]{organization={Department of Business Organisation, Universitat Polit\`ecnica de Val\`encia},
  city={Valencia},
  country={Spain}}

\begin{abstract}
Hybrid language models that interleave attention with recurrent components are increasingly competitive with pure Transformers, yet standard LoRA practice applies adapters uniformly without considering the distinct functional roles of each component type. We systematically study component-type LoRA placement across two hybrid architectures---\modelqwen (sequential, GatedDeltaNet + softmax attention) and \modelfalcon (parallel, Mamba-2 SSM + attention)---fine-tuned on three domains and evaluated on five benchmarks. We find that the attention pathway---despite being the minority component---consistently outperforms full-model adaptation with 5--10$\times$ fewer trainable parameters. Crucially, adapting the recurrent backbone is destructive in sequential hybrids ($-$14.8\pp on GSM8K) but constructive in parallel ones (+8.6\pp). We further document a transfer asymmetry: parallel hybrids exhibit positive cross-task transfer while sequential hybrids suffer catastrophic forgetting. These results establish that hybrid topology fundamentally determines adaptation response, and that component-aware LoRA placement is a necessary design dimension for hybrid architectures.
\end{abstract}


\end{frontmatter}

\section{Introduction}
\label{sec:introduction}

The past year has seen a proliferation of hybrid language models that combine softmax attention with alternative sequence-mixing mechanisms such as state-space models (SSMs) \cite{gu2023mamba,dao2024mamba2} or gated linear attention (GLA) \cite{yang2024gla}. Production examples include Jamba \cite{lieber2024jamba}, Zamba \cite{glorioso2024zamba}, Falcon-H1 \cite{zuo2025falconh1}, and Qwen3 \cite{qwen2025qwen3}, all of which achieve competitive or superior performance to pure Transformers while offering sub-quadratic inference cost. As these models enter practical deployment, practitioners naturally turn to Low-Rank Adaptation (LoRA) \cite{hu2021lora} for domain-specific fine-tuning. However, a fundamental question remains unaddressed: \emph{given that hybrid models contain structurally and functionally distinct component types, which components should LoRA target?}

Current LoRA practice overwhelmingly targets attention projections---a convention inherited from the pure-Transformer setting. Recent work has begun exploring PEFT for SSM-based models \cite{galim2025peft}, but the focus has been on pure-SSM architectures (Mamba, Mamba-2, Jamba) and individual SSM-internal parameters. No prior work has studied LoRA placement at the \emph{component-type} level in native hybrid architectures---that is, systematically comparing whether to adapt the attention pathway, the recurrent backbone, the MLP, or combinations thereof.

This question is particularly timely in light of recent evidence that the components of hybrid models serve asymmetric functional roles. Borobia \etal \cite{borobia2026functional} demonstrated via functional component ablation that in both sequential and parallel hybrids, the recurrent component acts as the ``backbone'' of the model (its removal causes perplexity degradation exceeding 35{,}000$\times$), while attention functions as a ``refinement'' mechanism (degradation of approximately 82$\times$). If the components play fundamentally different roles during inference, it is reasonable to hypothesize that they will also respond differently to adaptation.

In this paper, we present a controlled empirical study of component-type LoRA placement across two representative hybrid architectures:
\begin{itemize}[nosep]
  \item \textbf{\modelqwen} (sequential hybrid): 24 layers with 18 GatedDeltaNet/linear-attention layers and 6 full softmax attention layers in a 3:1 ratio.
  \item \textbf{\modelfalcon} (parallel hybrid): 36 blocks, each containing a softmax attention branch and a Mamba-2 SSM branch operating in parallel.
\end{itemize}

For each model, we define six placement conditions that isolate or combine the three component types (attention, recurrent backbone, MLP), fine-tune on three distinct domains (mathematics, code, general instruction), and evaluate across five standard benchmarks. Our contributions are as follows:

\begin{enumerate}[nosep]
  \item We introduce \textbf{component-type placement} as a dimension of LoRA configuration and provide the first systematic comparison across two hybrid topologies.
  \item We show that \textbf{attention-only placement} matches or exceeds full-model adaptation with 5--10$\times$ fewer parameters in both architectures, establishing it as the recommended default for hybrid LM fine-tuning.
  \item We demonstrate that \textbf{adapting the recurrent backbone is destructive in sequential hybrids but safe in parallel ones}, revealing the topology as the key determinant of adaptation response.
  \item We document a \textbf{transfer asymmetry}: parallel hybrids exhibit positive cross-task transfer where sequential hybrids show catastrophic forgetting, with attention-only placement providing the best protection in both cases.
\end{enumerate}

\section{Related Work}
\label{sec:related}

\subsection{Parameter-Efficient Fine-Tuning for Sequence Models}

LoRA \cite{hu2021lora} introduced trainable low-rank decomposition matrices alongside frozen pre-trained weights. Originally designed for Transformer attention projections, LoRA has become the dominant PEFT method for large language models. Subsequent variants (AdaLoRA \cite{zhang2023adalora}, DoRA \cite{liu2024dora}) refine the rank allocation or weight decomposition but retain the fundamental assumption that all target modules are architecturally homogeneous.

\subsection{PEFT for State-Space and Recurrent Models}

The emergence of SSM-based language models prompted investigation into which internal parameters are best suited for PEFT. Galim \etal \cite{galim2025peft} benchmarked LoRA, prefix tuning, and their proposed Selective Diagonal Tuning (SDT) across pure-SSM architectures (Mamba, Mamba-2, Jamba). Their key finding---that LoRA applied to linear projections outperforms LoRA on SSM-specific modules---is consistent with our results but does not extend to GatedDeltaNet or to parallel hybrid topologies. MambaPEFT \cite{greenewald2025mambapeft} at ICLR 2025 explored similar territory for Mamba-1/2.

Concurrently, Young \cite{young2026s0tuning} proposed S$_0$ Tuning, which adapts only the initial recurrent state $S_0$ in Qwen3.5 and Falcon-H1. This approach is complementary to weight-based LoRA: at 4B+ parameters, S$_0$ tuning is highly competitive, but at the sub-1B scale studied here, gains are marginal (+2.6 $\pm$ 3.7\pp, not statistically significant). Our work studies \emph{which weight matrices} to adapt rather than whether to adapt weights versus states.

\subsection{Hybrid Architecture Analysis}

Wang \etal \cite{wang2025hybrid} provided a systematic analysis of hybrid linear-attention Transformers, characterizing the design space along dimensions of layer ratio, placement pattern, and attention mechanism. Borobia \etal \cite{borobia2026functional} extended this via functional component ablation, establishing that both component types are essential and neither is bypassed, that the alternative component (GDN/SSM) serves as the primary language-modeling backbone, and that component importance follows a positional gradient with early layers being disproportionately critical. Our work builds directly on these functional insights to derive practical adaptation guidance.

\subsection{LoRA Targeting Strategies}

Prior work on \emph{where} to apply LoRA has focused on \emph{layer-level} selection: which layers of a homogeneous stack should receive adapters. Examples include importance-based layer selection, adaptive rank allocation (AdaLoRA \cite{zhang2023adalora}), and dynamic rank distribution methods. All of these operate within a single component type (typically attention). Our work introduces the orthogonal dimension of \emph{component-type} selection, asking not ``which layers'' but ``which kinds of modules.''

\section{Experimental Setup}
\label{sec:setup}

\subsection{Models}
\label{sec:models}

\begin{figure}[t]
  \centering
  \begin{subfigure}[b]{0.48\textwidth}
    \includegraphics[width=\textwidth]{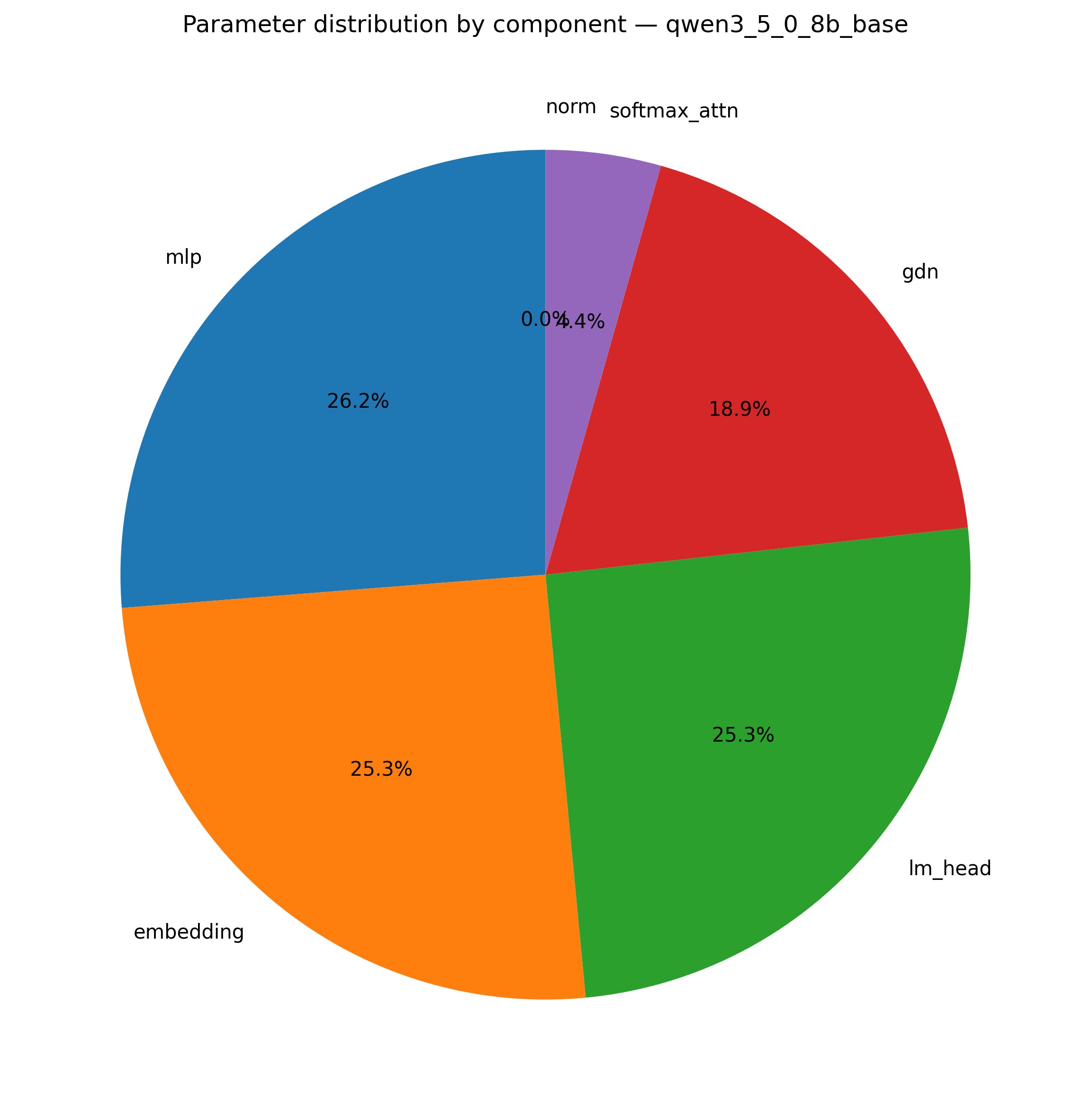}
    \caption{\modelqwen}
    \label{fig:pie_qwen}
  \end{subfigure}
  \hfill
  \begin{subfigure}[b]{0.48\textwidth}
    \includegraphics[width=\textwidth]{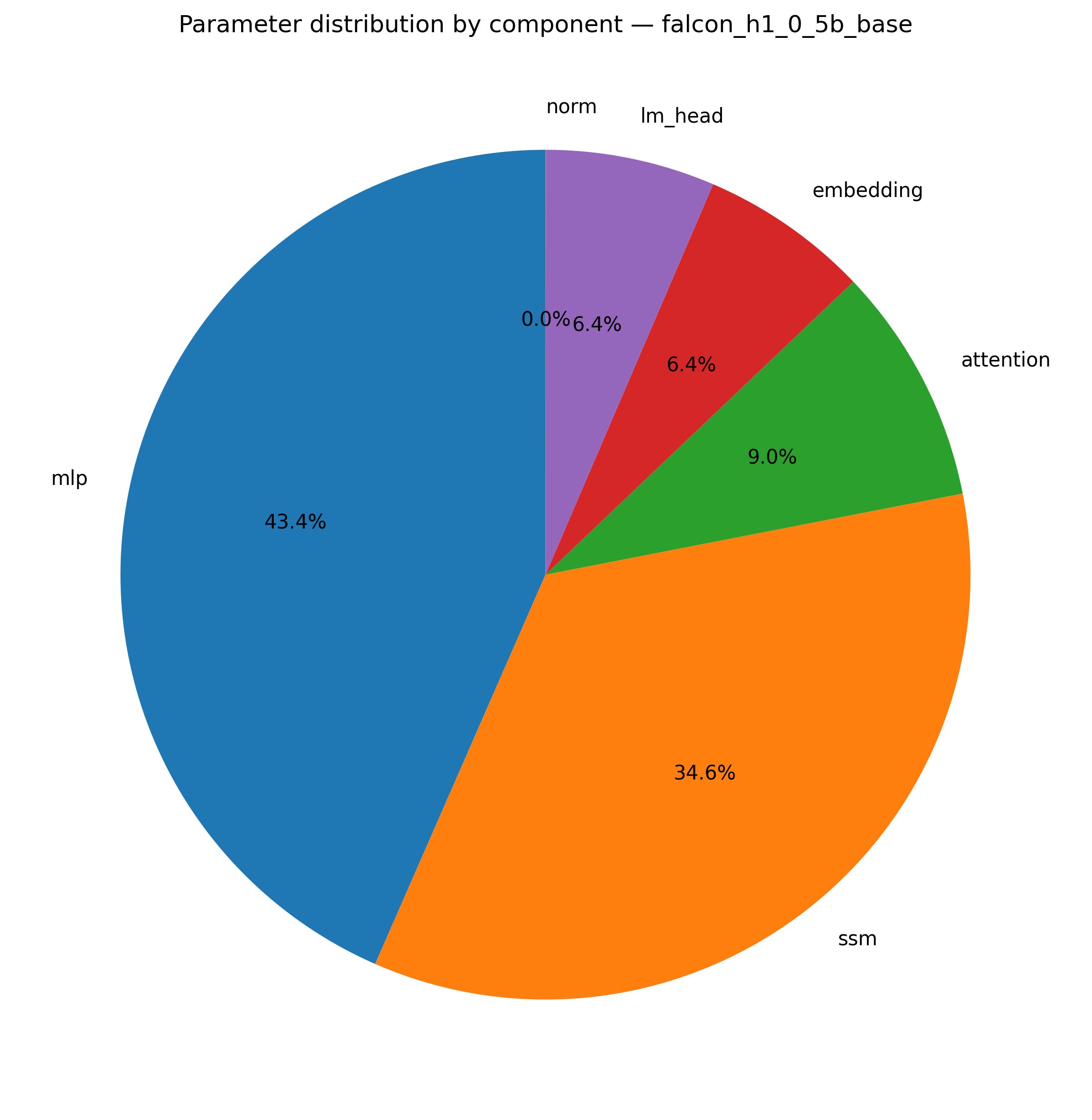}
    \caption{\modelfalcon}
    \label{fig:pie_falcon}
  \end{subfigure}
  \caption{Parameter distribution by component type. In both architectures, the attention mechanism constitutes a small fraction of total parameters (4.4\% for Qwen, 6.4\% for Falcon), yet proves to be the most effective LoRA target.}
  \label{fig:param_distribution}
\end{figure}

We select two sub-1B hybrid language models that represent the two dominant hybrid topologies:

\paragraph{Qwen3.5-0.8B-Base} \cite{qwen2025qwen3} is a sequential hybrid with 759M parameters organized into 24 layers. Of these, 18 use GatedDeltaNet (GDN) \cite{yang2024gla} as the sequence-mixing mechanism with linear attention, while 6 use full softmax attention, yielding a 3:1 GDN-to-attention ratio. The GDN component constitutes approximately 18.8\% of model parameters, softmax attention only 4.4\%, and the MLP 26.2\% (\Cref{fig:param_distribution}).

\paragraph{Falcon-H1-0.5B-Base} \cite{zuo2025falconh1} is a parallel hybrid with 524M parameters organized into 36 blocks. Each block contains a softmax attention branch and a Mamba-2 SSM \cite{dao2024mamba2} branch operating in parallel, with their outputs summed before the MLP. The SSM constitutes 34.6\% of parameters, attention 6.4\%, and the MLP 43.4\%.

Both architectures feature attention as a small minority of the parameter budget, with the recurrent component and MLP dominating.

\subsection{Module Discovery and Verification}
\label{sec:discovery}

To ensure reproducibility and correctness, we implemented an automated module discovery pipeline rather than relying on manual inspection. The pipeline operates in four stages: (1) load the model and traverse the full module tree; (2) classify each module into a component type (attention, recurrent, MLP, norm, embedding) based on its position in the hierarchy and naming conventions; (3) construct exact dotted-path target lists for each placement condition; and (4) verify the resulting PEFT model by confirming that all expected modules received adapters and no forbidden modules were modified.

For Falcon-H1, we exclude \texttt{conv1d} and \texttt{out\_proj} modules from SSM adaptation, following the constraints identified in the PEFT library and confirmed by preliminary experiments showing training instability when these modules are included. All 12 conditions (6 per model) passed verification with zero missing or unexpected adapter hosts.

\subsection{Placement Conditions}
\label{sec:conditions}

For each model, we define six placement conditions that systematically isolate and combine the three component types. \Cref{tab:conditions} summarizes the conditions and their parameter budgets.

\begin{table}[t]
\centering
\caption{Placement conditions and trainable parameter counts at rank $r{=}16$. ``Attn'' refers to softmax attention; ``Rec'' refers to the recurrent backbone (GDN for Qwen, SSM for Falcon).}
\label{tab:conditions}
\small
\begin{tabular}{@{}llrrr@{}}
\toprule
\textbf{Model} & \textbf{Condition} & \textbf{Params (M)} & \textbf{\% Model} & \textbf{Modules} \\
\midrule
\multirow{6}{*}{\modelqwen}
  & \texttt{all\_layers}        & 10.82 & 1.42\% & 186 \\
  & \texttt{softmax\_only}      &  1.08 & 0.14\% &  24 \\
  & \texttt{gdn\_only}          &  4.43 & 0.59\% &  90 \\
  & \texttt{mlp\_only}          &  5.31 & 0.70\% &  72 \\
  & \texttt{softmax\_plus\_mlp} &  6.39 & 0.84\% &  96 \\
  & \texttt{gdn\_plus\_mlp}     &  9.74 & 1.28\% & 162 \\
\midrule
\multirow{6}{*}{\modelfalcon}
  & \texttt{all\_eligible}         & 11.47 & 2.15\% & 289 \\
  & \texttt{attention\_only}       &  2.21 & 0.42\% & 108 \\
  & \texttt{ssm\_only}             &  2.52 & 0.48\% &  36 \\
  & \texttt{mlp\_only}             &  5.31 & 1.01\% & 108 \\
  & \texttt{attention\_plus\_mlp}  &  7.52 & 1.42\% & 216 \\
  & \texttt{ssm\_plus\_mlp}        &  7.83 & 1.48\% & 144 \\
\bottomrule
\end{tabular}
\end{table}

\subsection{Fine-Tuning Domains}
\label{sec:domains}

We select three training domains representing distinct capability profiles:

\begin{itemize}[nosep]
  \item \textbf{GSM8K} \cite{cobbe2021gsm8k}: 2{,}000 samples of grade-school mathematics, targeting numerical reasoning.
  \item \textbf{CodeAlpaca} \cite{codealpaca2023}: 2{,}000 samples of code instruction-following, targeting code generation.
  \item \textbf{UltraChat} \cite{ding2023ultrachat}: 2{,}000 samples of general conversational instruction, targeting broad instruction-following.
\end{itemize}

The choice of three semantically diverse domains enables cross-task evaluation: we train on one domain and evaluate on all benchmarks, allowing us to measure both target-task improvement and off-target forgetting or transfer.

\subsection{Evaluation Protocol}
\label{sec:eval}

We evaluate all conditions on five benchmarks:
\begin{itemize}[nosep]
  \item \textbf{MMLU} \cite{hendrycks2021mmlu}: 512 samples, measuring broad knowledge.
  \item \textbf{GSM8K} \cite{cobbe2021gsm8k}: 128--256 samples, measuring mathematical reasoning.
  \item \textbf{ARC-Challenge} \cite{clark2018arc}: 299--512 samples, measuring scientific reasoning.
  \item \textbf{HellaSwag} \cite{zellers2019hellaswag}: 512 samples, measuring commonsense reasoning.
  \item \textbf{HumanEval} \cite{chen2021humaneval}: 164 samples, measuring code generation (pass@1).
\end{itemize}

At the sub-1B scale, no placement condition on either model produced functionally correct code on HumanEval (pass@1 $\leq$ 0.6\% across all conditions). We therefore exclude HumanEval from quantitative comparisons and note this as a scale limitation. All evaluations use greedy decoding to ensure deterministic comparison.

\subsection{Training Configuration}
\label{sec:training_config}

All experiments use identical hyperparameters: LoRA rank $r{=}16$, $\alpha{=}32$ (scaling factor 2.0), dropout 0.05, learning rate $2{\times}10^{-4}$ with cosine schedule, 3 epochs, effective batch size 16 (batch size 4, gradient accumulation 4), maximum sequence length 1{,}024, 8-bit Adam optimizer, bf16 mixed precision, and gradient checkpointing. Hardware consisted of NVIDIA L4 24\,GB (Google Colab Pro) and RTX 4090 24\,GB (RunPod). A single training run requires approximately 15--40 minutes depending on the model and hardware.

\section{Results}
\label{sec:results}

\subsection{Baselines}
\label{sec:baselines}

\Cref{tab:baselines} reports the pre-training baselines for both models. Despite its smaller parameter count, \modelfalcon achieves higher MMLU and GSM8K scores than \modelqwen, while \modelqwen leads on ARC-Challenge and HellaSwag.

\begin{table}[h]
\centering
\caption{Baseline accuracy (no fine-tuning) on four benchmarks.}
\label{tab:baselines}
\small
\begin{tabular}{@{}lcccc@{}}
\toprule
\textbf{Model} & \textbf{MMLU} & \textbf{GSM8K} & \textbf{ARC-C} & \textbf{HellaSwag} \\
\midrule
\modelqwen  & 0.482 & 0.297 & 0.732 & 0.416 \\
\modelfalcon & 0.518 & 0.383 & 0.743 & 0.330 \\
\bottomrule
\end{tabular}
\end{table}

\subsection{Component-Specific Placement: Main Results}
\label{sec:main_results}

\Cref{tab:main_gsm8k} presents the core results for GSM8K-trained models, the domain where placement effects are most pronounced. Complete results for all domain--benchmark combinations are provided in \Cref{tab:full_qwen,tab:full_falcon}.

\begin{table}[t]
\centering
\caption{Benchmark accuracy after fine-tuning on GSM8K. \textbf{Bold} marks the best condition per model. $\Delta_\text{base}$ is the change in GSM8K accuracy relative to the pre-training baseline (percentage points). Conditions are ordered by GSM8K performance.}
\label{tab:main_gsm8k}
\small
\begin{tabular}{@{}llccccr@{}}
\toprule
\textbf{Model} & \textbf{Condition} & \textbf{MMLU} & \textbf{GSM8K} & \textbf{ARC-C} & \textbf{HSwag} & \textbf{$\Delta_\text{base}$} \\
\midrule
\multirow{6}{*}{\rotatebox[origin=c]{90}{\footnotesize \modelqwen}}
  & \texttt{softmax\_only}       & .490 & \textbf{.398} & .726 & .398 & +10.2 \\
  & \texttt{mlp\_only}           & .477 & .383 & .709 & .459 & +8.6  \\
  & \texttt{all\_layers}         & .473 & .375 & .746 & .434 & +7.8  \\
  & \texttt{gdn\_plus\_mlp}      & .502 & .203 & .712 & .434 & $-$9.4  \\
  & \texttt{gdn\_only}           & .488 & .148 & .732 & .451 & $-$14.8 \\
  & \texttt{softmax\_plus\_mlp}  & .481 & .148 & .706 & .416 & $-$14.8 \\
\midrule
\multirow{6}{*}{\rotatebox[origin=c]{90}{\footnotesize \modelfalcon}}
  & \texttt{attention\_only}      & .502 & \textbf{.555} & .732 & .320 & +17.2 \\
  & \texttt{mlp\_only}            & .506 & .508 & .709 & .326 & +12.5 \\
  & \texttt{ssm\_plus\_mlp}       & .516 & .508 & .712 & .307 & +12.5 \\
  & \texttt{all\_eligible}        & .516 & .504 & .712 & .303 & +12.1 \\
  & \texttt{attention\_plus\_mlp} & .510 & .492 & .709 & .326 & +10.9 \\
  & \texttt{ssm\_only}            & .523 & .469 & .729 & .305 & +8.6  \\
\bottomrule
\end{tabular}
\end{table}

Several patterns emerge. First, attention-only placement achieves the highest GSM8K accuracy in both models---with the fewest trainable parameters. In \modelqwen, \texttt{softmax\_only} reaches 0.398 with only 1.08M parameters (0.14\% of the model), outperforming \texttt{all\_layers} (0.375) which uses 10$\times$ more parameters. In \modelfalcon, \texttt{attention\_only} reaches 0.555 with 2.21M parameters (0.42\%), again surpassing \texttt{all\_eligible} (0.504) at 5.2$\times$ the parameter count.

Second, the recurrent backbone responds strikingly differently depending on the topology. In the sequential \modelqwen, \texttt{gdn\_only} \emph{destroys} GSM8K performance ($-$14.8\pp below baseline), making it worse than the un-fine-tuned model. In the parallel \modelfalcon, \texttt{ssm\_only} \emph{improves} GSM8K by +8.6\pp. This topology-dependent asymmetry is the central finding of this paper.

Third, an unexpected destructive interference pattern appears: in \modelqwen, \texttt{softmax\_only} (0.398) and \texttt{mlp\_only} (0.383) both improve GSM8K, yet their union \texttt{softmax\_plus\_mlp} (0.148) is dramatically worse---a 25\pp collapse below either component alone. This anomaly is not a training failure: MMLU, ARC-Challenge, and HellaSwag remain normal for \texttt{softmax\_plus\_mlp}, and the pattern is consistent across all three training domains (\Cref{tab:full_qwen,tab:full_falcon}).

\begin{figure}[t]
  \centering
  \begin{subfigure}[b]{0.48\textwidth}
    \includegraphics[width=\textwidth]{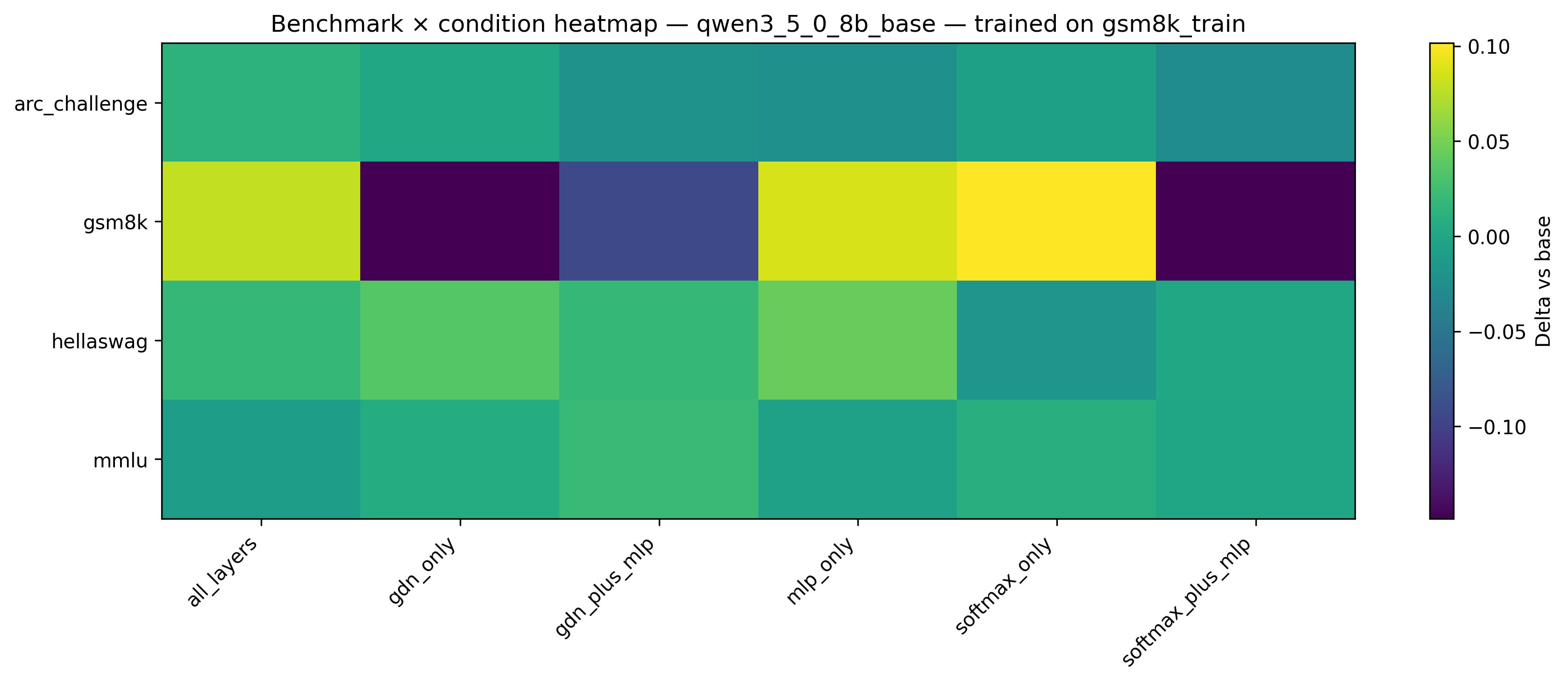}
    \caption{\modelqwen trained on GSM8K}
    \label{fig:heatmap_qwen_gsm8k}
  \end{subfigure}
  \hfill
  \begin{subfigure}[b]{0.48\textwidth}
    \includegraphics[width=\textwidth]{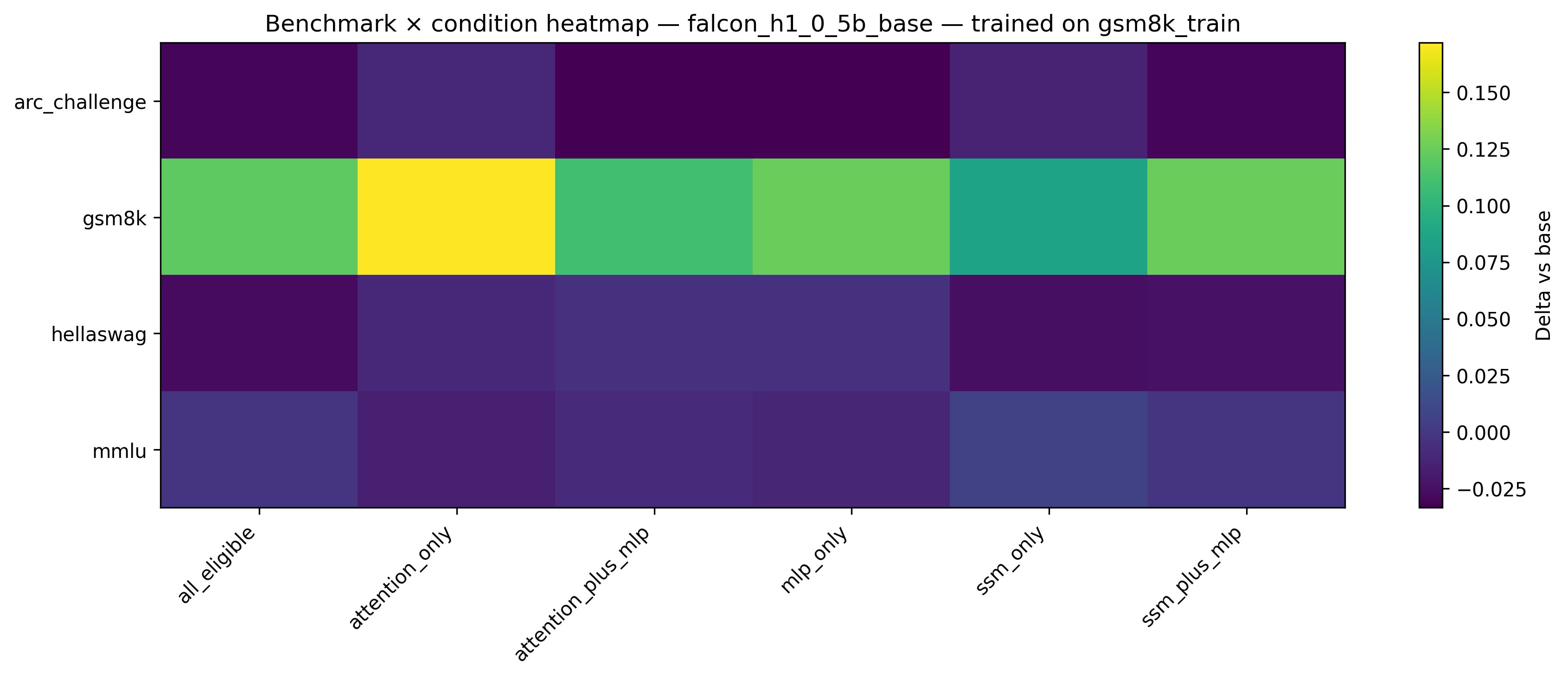}
    \caption{\modelfalcon trained on GSM8K}
    \label{fig:heatmap_falcon_gsm8k}
  \end{subfigure}
  \caption{Benchmark $\times$ condition heatmaps (delta versus base accuracy). Warm colors indicate improvement, cool colors degradation. Falcon shows uniformly positive GSM8K gains across all conditions, while Qwen exhibits severe degradation for GDN-involved and \texttt{softmax\_plus\_mlp} conditions.}
  \label{fig:heatmap_gsm8k}
\end{figure}

\subsection{Cross-Task Transfer and Forgetting}
\label{sec:transfer}

Fine-tuning on one domain inevitably affects performance on others. We quantify this via the \emph{forgetting score}, defined as the mean accuracy change on non-target benchmarks relative to baseline. A positive forgetting score indicates degradation on off-target tasks; a negative score indicates positive transfer.

\begin{figure}[t]
  \centering
  \begin{subfigure}[b]{0.48\textwidth}
    \includegraphics[width=\textwidth]{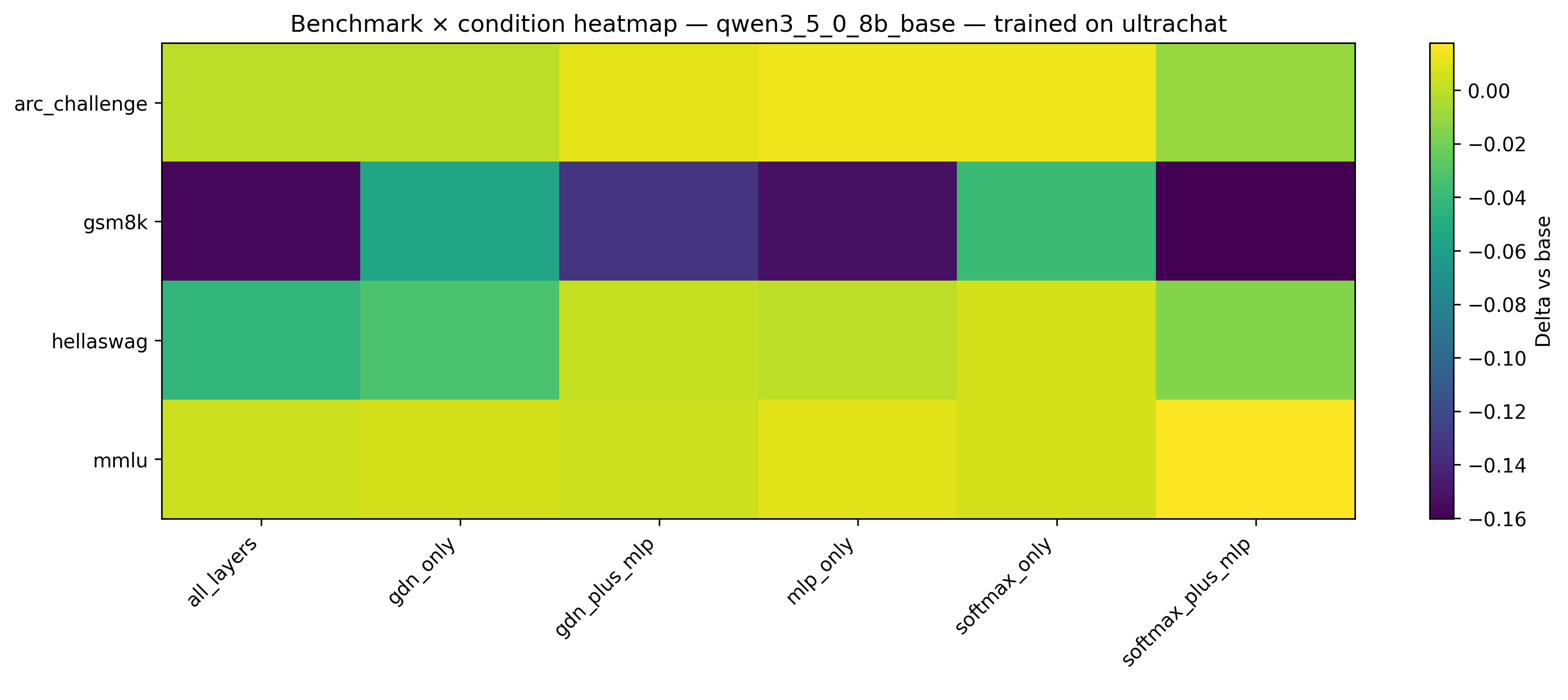}
    \caption{\modelqwen trained on UltraChat}
    \label{fig:heatmap_qwen_ultrachat}
  \end{subfigure}
  \hfill
  \begin{subfigure}[b]{0.48\textwidth}
    \includegraphics[width=\textwidth]{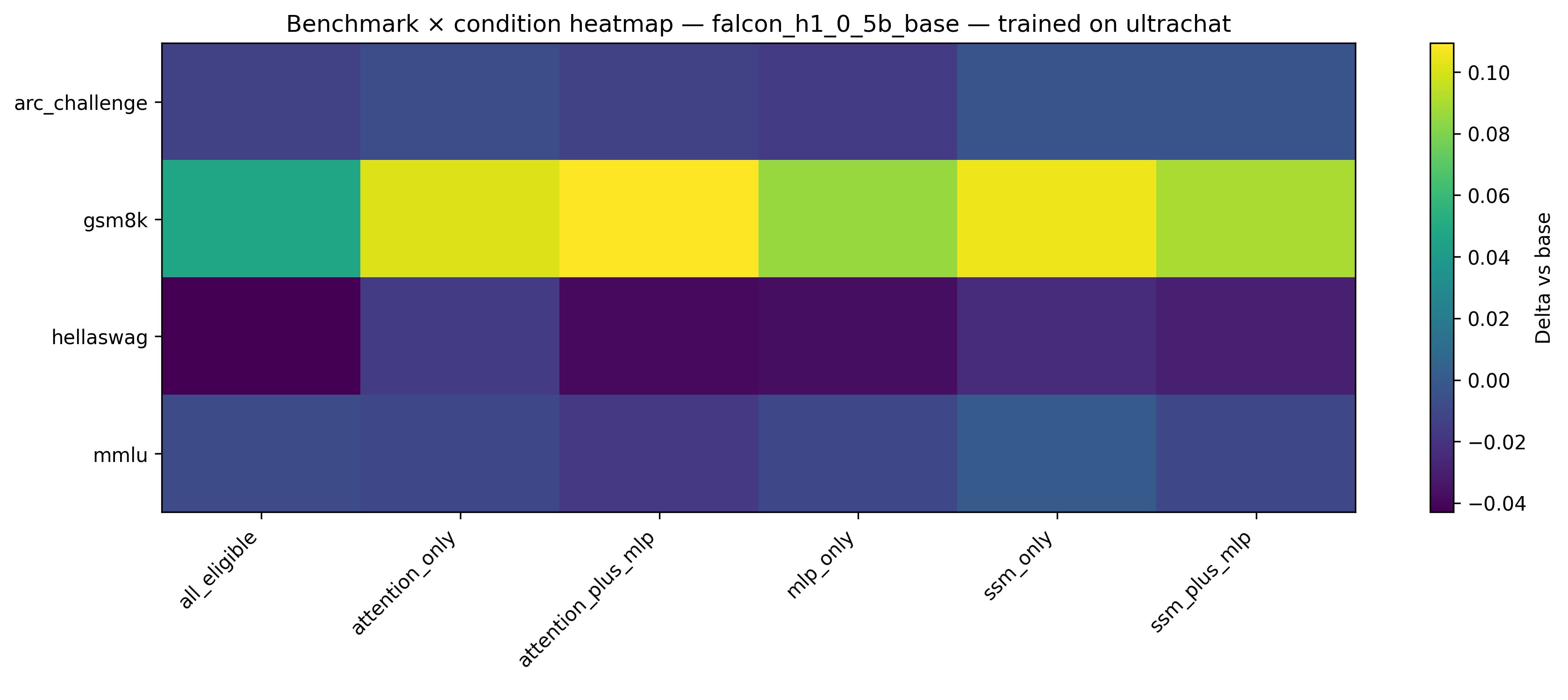}
    \caption{\modelfalcon trained on UltraChat}
    \label{fig:heatmap_falcon_ultrachat}
  \end{subfigure}
  \caption{Cross-task transfer after UltraChat training. Qwen shows pervasive forgetting (cool tones, especially on GSM8K), while Falcon shows positive transfer on GSM8K (warm tones). The contrast is most visible in the GSM8K row.}
  \label{fig:heatmap_ultrachat}
\end{figure}

\Cref{fig:heatmap_ultrachat} reveals a dramatic asymmetry. After training on UltraChat (general instruction), \modelqwen loses up to 16.0\pp on GSM8K---a task it was never trained on---with the most severe forgetting in \texttt{softmax\_plus\_mlp} ($-$16.0\pp) and \texttt{all\_layers} ($-$15.6\pp). In contrast, \modelfalcon \emph{gains} on GSM8K after UltraChat training, with improvements of up to +10.9\pp (\texttt{attention\_plus\_mlp}) and +10.5\pp (\texttt{ssm\_only}). This positive transfer suggests that the parallel topology is fundamentally more robust: general instruction tuning ``lifts all boats'' rather than overwriting domain-specific capabilities.

Attention-only placement provides the best forgetting protection across both architectures. In the CodeAlpaca $\rightarrow$ GSM8K setting (code training, math evaluation), Falcon's \texttt{attention\_only} shows near-zero forgetting (+0.8\pp on GSM8K), while \texttt{all\_eligible} degrades by $-$2.3\pp. For Qwen, \texttt{softmax\_only} loses $-$11.3\pp on GSM8K after code training---substantial, but less than \texttt{all\_layers} at $-$21.9\pp.

\subsection{Efficiency Analysis}
\label{sec:efficiency}

\begin{figure}[t]
  \centering
  \begin{subfigure}[b]{0.48\textwidth}
    \includegraphics[width=\textwidth]{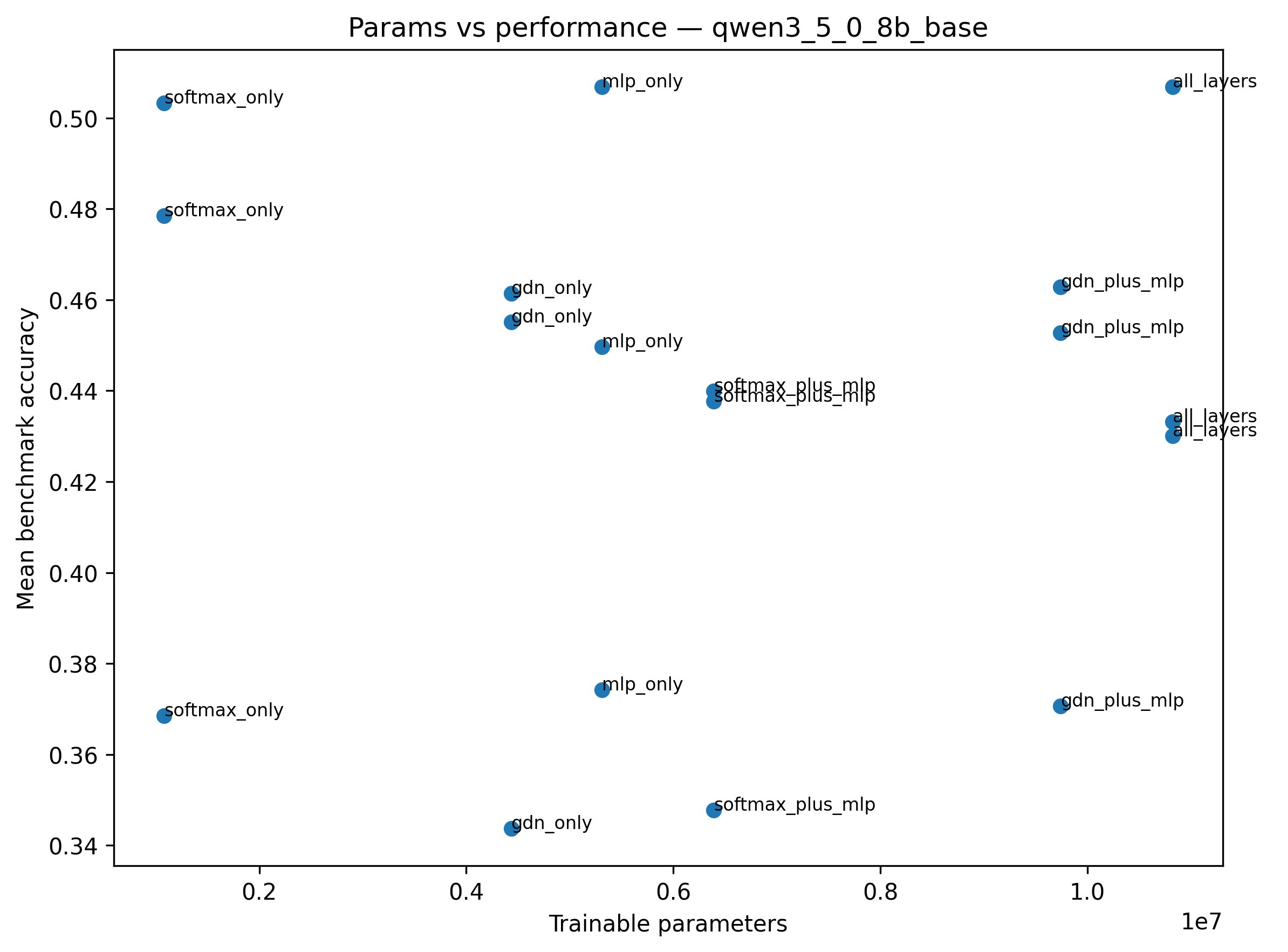}
    \caption{\modelqwen}
    \label{fig:pareto_qwen}
  \end{subfigure}
  \hfill
  \begin{subfigure}[b]{0.48\textwidth}
    \includegraphics[width=\textwidth]{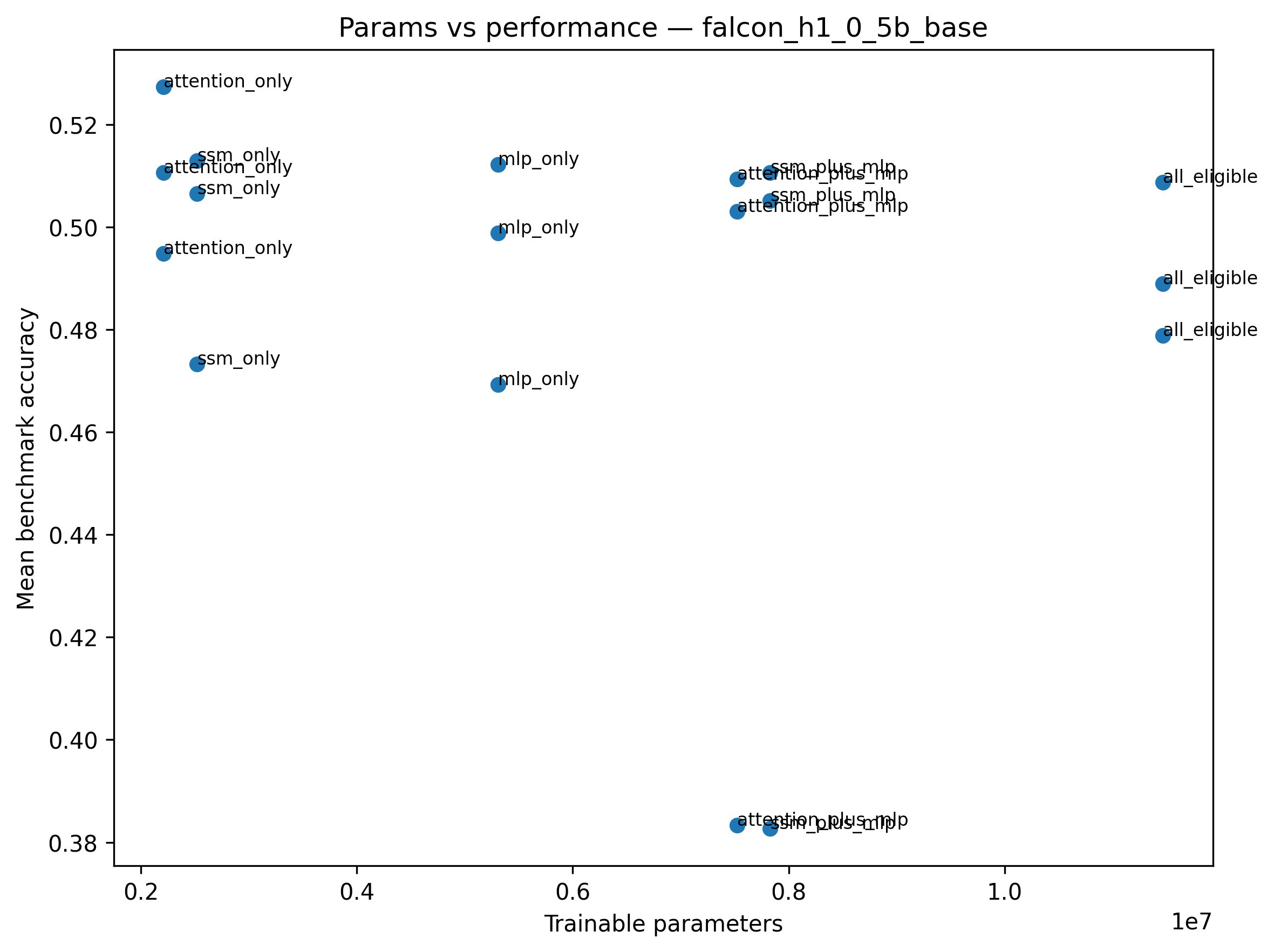}
    \caption{\modelfalcon}
    \label{fig:pareto_falcon}
  \end{subfigure}
  \caption{Trainable parameters versus mean benchmark accuracy (Pareto scatter). Each point is one (condition, training domain) pair. Attention-only placements (upper left in both plots) dominate the Pareto frontier: highest accuracy at lowest parameter cost.}
  \label{fig:pareto}
\end{figure}

The parameter efficiency of attention-only placement is striking. \Cref{fig:pareto} shows the Pareto frontier of trainable parameters versus mean benchmark accuracy across all training domains. In both models, the attention-only condition consistently achieves the highest or near-highest mean accuracy at the lowest parameter cost, dominating the upper-left region of the Pareto plot.

For GSM8K-trained Qwen, \texttt{softmax\_only} achieves a target-task efficiency of 9.4\pp per million parameters versus 0.7\pp/M for \texttt{all\_layers}---a 13$\times$ advantage. For Falcon, \texttt{attention\_only} achieves 7.8\pp/M versus 1.1\pp/M for \texttt{all\_eligible}---a 7$\times$ advantage. These efficiency ratios are summarized in \Cref{tab:efficiency}.

\begin{table}[t]
\centering
\caption{Efficiency comparison for GSM8K-trained conditions. Efficiency ratio is defined as target-task accuracy gain (pp) per million trainable parameters.}
\label{tab:efficiency}
\small
\begin{tabular}{@{}llrrr@{}}
\toprule
\textbf{Model} & \textbf{Condition} & \textbf{Params (M)} & \textbf{$\Delta$GSM8K (pp)} & \textbf{pp/M} \\
\midrule
\multirow{3}{*}{\modelqwen}
  & \texttt{softmax\_only} & 1.08 & +10.2 & 9.4 \\
  & \texttt{mlp\_only}     & 5.31 & +8.6  & 1.6 \\
  & \texttt{all\_layers}   & 10.82 & +7.8 & 0.7 \\
\midrule
\multirow{3}{*}{\modelfalcon}
  & \texttt{attention\_only} & 2.21 & +17.2 & 7.8 \\
  & \texttt{mlp\_only}       & 5.31 & +12.5 & 2.4 \\
  & \texttt{all\_eligible}   & 11.47 & +12.1 & 1.1 \\
\bottomrule
\end{tabular}
\end{table}

\subsection{Radar Profiles}
\label{sec:radar}

\begin{figure}[t]
  \centering
  \begin{subfigure}[b]{0.48\textwidth}
    \includegraphics[width=\textwidth]{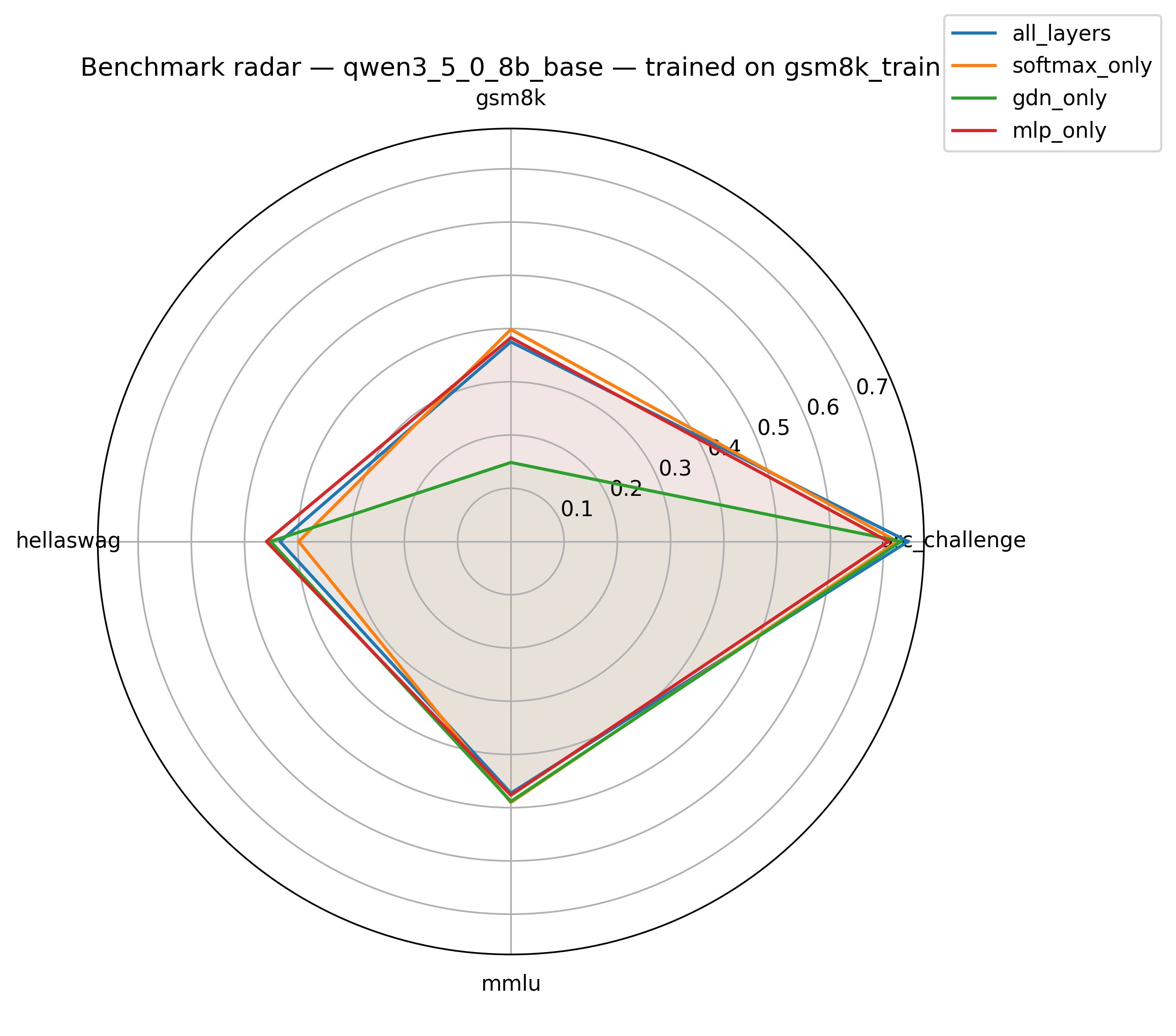}
    \caption{\modelqwen trained on GSM8K}
    \label{fig:radar_qwen_gsm8k}
  \end{subfigure}
  \hfill
  \begin{subfigure}[b]{0.48\textwidth}
    \includegraphics[width=\textwidth]{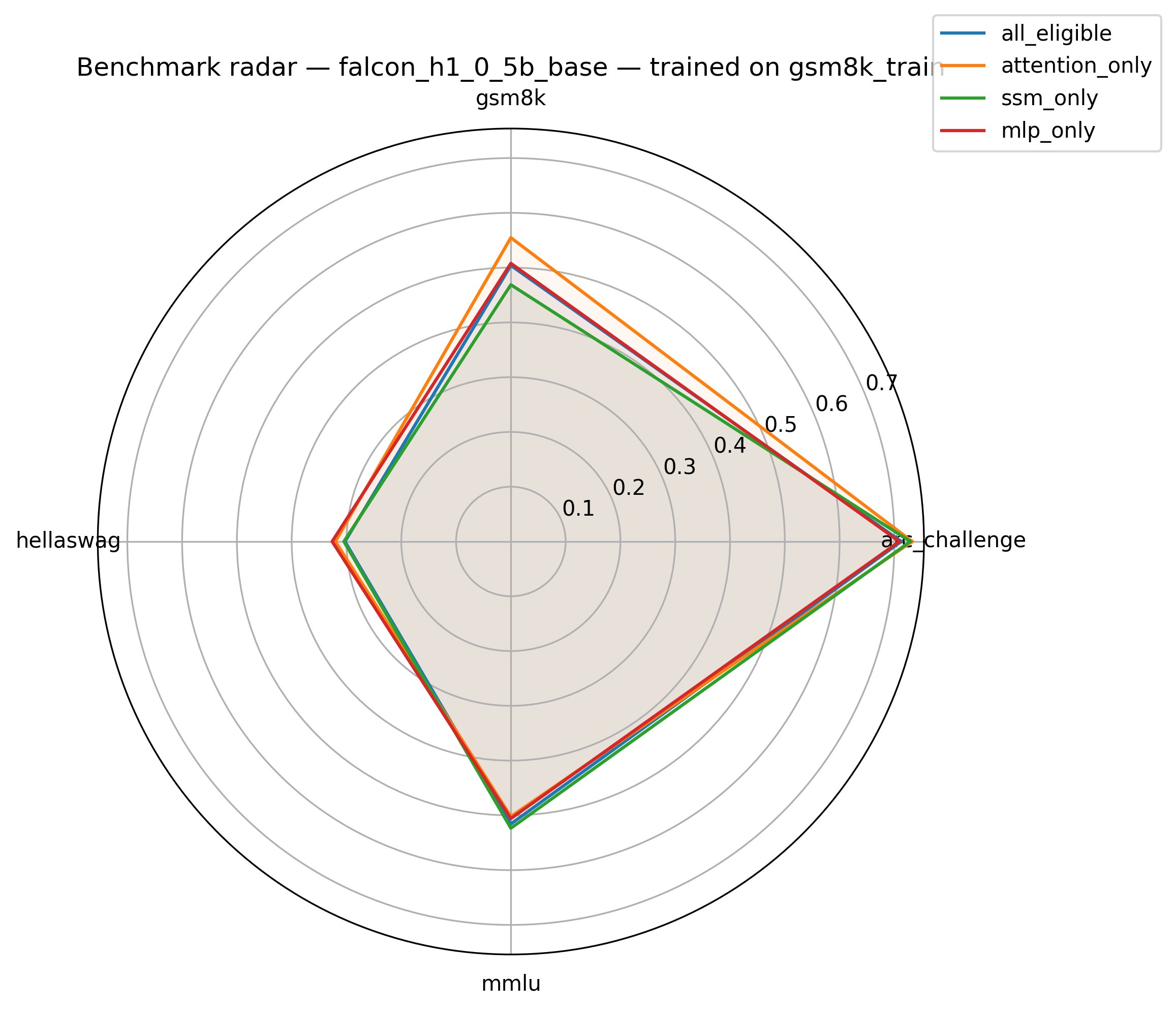}
    \caption{\modelfalcon trained on GSM8K}
    \label{fig:radar_falcon_gsm8k}
  \end{subfigure}
  \caption{Radar chart comparison of single-component conditions after GSM8K training. Four axes correspond to the four benchmarks (excluding HumanEval). Attention-only (orange) dominates the GSM8K axis in both models.}
  \label{fig:radar_gsm8k}
\end{figure}

\Cref{fig:radar_gsm8k} presents radar charts comparing the four single-component conditions after GSM8K training. The Falcon radar (\Cref{fig:radar_falcon_gsm8k}) shows that \texttt{attention\_only} dominates on GSM8K while maintaining competitive ARC-Challenge and MMLU, confirming its superiority is not limited to the target benchmark. The Qwen radar (\Cref{fig:radar_qwen_gsm8k}) reveals a more differentiated profile: \texttt{softmax\_only} leads on GSM8K but trades off HellaSwag, while \texttt{mlp\_only} exhibits the opposite pattern.

\subsection{Statistical Validation}
\label{sec:stats}

We computed paired bootstrap confidence intervals for key comparisons using the UltraChat-trained conditions (the only setting where per-instance paired data was available across all benchmarks). For Falcon trained on UltraChat:
\begin{itemize}[nosep]
  \item \texttt{attention\_only} vs.\ \texttt{all\_eligible} on HellaSwag: mean difference +2.7\pp, 95\% CI [+0.8, +4.7], significant.
  \item \texttt{attention\_only} vs.\ \texttt{all\_eligible} on GSM8K: mean difference +5.5\pp, 95\% CI [$-$1.2, +12.9], not significant at $\alpha{=}0.05$ but trending.
  \item \texttt{ssm\_only} vs.\ \texttt{attention\_only} on all benchmarks: CIs include zero, indicating comparable performance.
\end{itemize}

The single-seed design (see \Cref{sec:limitations}) limits the statistical power of these comparisons. However, the large effect sizes observed---particularly the 17.2\pp GSM8K advantage of Falcon's \texttt{attention\_only} and the 14.8\pp destructive effect of Qwen's \texttt{gdn\_only}---greatly exceed typical run-to-run variance in LoRA fine-tuning, supporting the reliability of the primary findings.

\section{Full Results}
\label{sec:full_results}

\Cref{tab:full_qwen,tab:full_falcon} report the complete accuracy matrix across all training domain $\times$ condition $\times$ benchmark combinations for each model.

\begin{table}[t]
\centering
\caption{Complete benchmark accuracy for \modelqwen across all placement conditions and training domains. Best result per column within each domain group is \textbf{bolded}. Baseline (no fine-tuning): MMLU\,=\,.482, GSM8K\,=\,.297, ARC-C\,=\,.732, HSwag\,=\,.416.}
\label{tab:full_qwen}
\small
\setlength{\tabcolsep}{4pt}
\begin{tabular}{@{}llcccc@{}}
\toprule
\textbf{Train Domain} & \textbf{Condition} & \textbf{MMLU} & \textbf{GSM8K} & \textbf{ARC-C} & \textbf{HSwag} \\
\midrule
\multirow{6}{*}{GSM8K}
  & \texttt{softmax\_only}      & .490 & \textbf{.398} & .726 & .398 \\
  & \texttt{mlp\_only}          & .477 & .383 & .709 & \textbf{.459} \\
  & \texttt{all\_layers}        & .473 & .375 & \textbf{.746} & .434 \\
  & \texttt{gdn\_plus\_mlp}     & \textbf{.502} & .203 & .712 & .434 \\
  & \texttt{gdn\_only}          & .488 & .148 & .732 & .451 \\
  & \texttt{softmax\_plus\_mlp} & .481 & .148 & .706 & .416 \\
\midrule
\multirow{6}{*}{CodeAlpaca}
  & \texttt{softmax\_only}      & \textbf{.506} & .184 & .719 & \textbf{.428} \\
  & \texttt{mlp\_only}          & .496 & .203 & \textbf{.746} & .426 \\
  & \texttt{gdn\_plus\_mlp}     & .479 & \textbf{.227} & .722 & .426 \\
  & \texttt{all\_layers}        & .486 & .078 & .722 & .434 \\
  & \texttt{gdn\_only}          & .479 & .102 & .736 & .397 \\
  & \texttt{softmax\_plus\_mlp} & .494 & .098 & .729 & .412 \\
\midrule
\multirow{6}{*}{UltraChat}
  & \texttt{softmax\_only}      & .488 & \textbf{.258} & \textbf{.746} & \textbf{.422} \\
  & \texttt{gdn\_only}          & .488 & .242 & .732 & .383 \\
  & \texttt{gdn\_plus\_mlp}     & .486 & .164 & .742 & .418 \\
  & \texttt{mlp\_only}          & .492 & .145 & .746 & .416 \\
  & \texttt{all\_layers}        & .486 & .141 & .732 & .373 \\
  & \texttt{softmax\_plus\_mlp} & \textbf{.500} & .137 & .722 & .400 \\
\bottomrule
\end{tabular}
\end{table}

\begin{table}[t]
\centering
\caption{Complete benchmark accuracy for \modelfalcon across all placement conditions and training domains. Best result per column within each domain group is \textbf{bolded}. Baseline (no fine-tuning): MMLU\,=\,.518, GSM8K\,=\,.383, ARC-C\,=\,.743, HSwag\,=\,.330.}
\label{tab:full_falcon}
\small
\setlength{\tabcolsep}{4pt}
\begin{tabular}{@{}llcccc@{}}
\toprule
\textbf{Train Domain} & \textbf{Condition} & \textbf{MMLU} & \textbf{GSM8K} & \textbf{ARC-C} & \textbf{HSwag} \\
\midrule
\multirow{6}{*}{GSM8K}
  & \texttt{attention\_only}     & .502 & \textbf{.555} & \textbf{.732} & .320 \\
  & \texttt{mlp\_only}           & .506 & .508 & .709 & \textbf{.326} \\
  & \texttt{ssm\_plus\_mlp}      & .516 & .508 & .712 & .307 \\
  & \texttt{all\_eligible}       & .516 & .504 & .712 & .303 \\
  & \texttt{attention\_plus\_mlp} & .510 & .492 & .709 & .326 \\
  & \texttt{ssm\_only}           & \textbf{.523} & .469 & .729 & .305 \\
\midrule
\multirow{6}{*}{CodeAlpaca}
  & \texttt{attention\_only}     & .514 & \textbf{.391} & \textbf{.739} & \textbf{.336} \\
  & \texttt{ssm\_plus\_mlp}      & \textbf{.520} & .352 & .726 & .316 \\
  & \texttt{attention\_plus\_mlp} & .506 & .363 & .729 & .318 \\
  & \texttt{all\_eligible}       & .516 & .359 & .726 & .314 \\
  & \texttt{ssm\_only}           & .518 & .320 & .729 & .326 \\
  & \texttt{mlp\_only}           & .512 & .320 & .732 & .313 \\
\midrule
\multirow{6}{*}{UltraChat}
  & \texttt{attention\_plus\_mlp} & .500 & \textbf{.492} & .729 & .291 \\
  & \texttt{ssm\_only}           & \textbf{.518} & .488 & \textbf{.739} & .307 \\
  & \texttt{attention\_only}     & .508 & .484 & .736 & \textbf{.314} \\
  & \texttt{ssm\_plus\_mlp}      & .508 & .473 & .739 & .301 \\
  & \texttt{mlp\_only}           & .508 & .469 & .726 & .293 \\
  & \texttt{all\_eligible}       & .510 & .430 & .729 & .287 \\
\bottomrule
\end{tabular}
\end{table}

\section{Discussion}
\label{sec:discussion}

\subsection{Why the Minority Component is the Best Target}
\label{sec:why_attention}

The most striking result is that the attention pathway---the minority component by parameter count in both architectures---is the most effective LoRA target. We propose this reflects a division of labor established during pre-training: the recurrent backbone (GDN or SSM) encodes a rigid sequential prior that is highly optimized and does not tolerate perturbation, while attention provides a more ``plastic'' refinement mechanism that can absorb task-specific adaptations without disrupting the model's core representations.

This interpretation is consistent with the functional ablation results of Borobia \etal \cite{borobia2026functional}, who found that removing the recurrent component causes catastrophic performance collapse (35{,}200$\times$ perplexity increase) while removing attention causes moderate degradation (82$\times$). A component that is ``load-bearing'' for forward-pass functionality may also be ``fragile'' under parameter perturbation---precisely because its weights encode tightly coupled sequential dynamics that low-rank updates can easily disrupt.

\subsection{Sequential vs.\ Parallel: The Topology Matters}
\label{sec:topology}

The topology of the hybrid architecture---sequential versus parallel---emerges as the key determinant of adaptation behavior. In the sequential \modelqwen, the GDN component processes the full residual stream before attention; perturbing GDN therefore disrupts the input to all downstream computations, including attention and MLP. In the parallel \modelfalcon, the SSM and attention branches process the \emph{same} input independently, and their outputs are summed; perturbing one branch does not corrupt the other's input. This architectural difference explains why GDN adaptation is destructive in Qwen but SSM adaptation is safe in Falcon.

This finding has immediate practical implications: for sequential hybrids, practitioners should \emph{avoid} adapting the recurrent backbone entirely, whereas for parallel hybrids, all components are viable targets (though attention-only remains most efficient).

\subsection{The Destructive Interference Anomaly}
\label{sec:anomaly}

The \texttt{softmax\_plus\_mlp} anomaly in \modelqwen merits discussion. When \texttt{softmax\_only} and \texttt{mlp\_only} each improve GSM8K individually (+10.2\pp and +8.6\pp respectively), their combination collapses to $-$14.8\pp below baseline. This is not explained by increased parameter count (the combined condition uses fewer parameters than \texttt{all\_layers}, which achieves +7.8\pp) and is specific to GSM8K---other benchmarks remain normal.

We hypothesize that this reflects a conflict in the optimization landscape: when both the refinement pathway (attention) and the transformation pathway (MLP) are simultaneously adapted in a sequential architecture where the backbone is frozen, gradient updates may push the two components toward incompatible representations. The backbone, being frozen, cannot mediate this conflict. In \texttt{all\_layers}, the GDN backbone \emph{can} co-adapt, which---despite the destructive effect of GDN adaptation on its own---paradoxically provides enough flexibility to prevent the interference between attention and MLP. This pattern is consistent across all three training domains (\texttt{softmax\_plus\_mlp} achieves 0.148 on GSM8K for the math domain, 0.098 for code, and 0.137 for general instruction), reinforcing that it is a structural phenomenon rather than a training artifact.

\subsection{Connection to Functional Roles}
\label{sec:functional_connection}

Our results directly extend the functional ablation framework of Borobia \etal \cite{borobia2026functional}. That work established a hierarchy of functional importance during \emph{inference}: backbone $>$ attention. We now show that this hierarchy is \emph{inverted} for \emph{adaptation}: attention $>$ backbone. The component most important for forward-pass computation is the least suitable for low-rank modification, and vice versa. This creates a coherent picture: the backbone is a rigid, optimized structure that the model relies on heavily and that does not tolerate perturbation; attention is a flexible, secondary mechanism that readily absorbs task-specific changes.

\subsection{Practical Recommendations}
\label{sec:recommendations}

Based on our findings, we propose the following practitioner guidelines for LoRA placement in hybrid language models:

\begin{enumerate}[nosep]
  \item \textbf{Default to attention-only placement.} This achieves competitive or superior accuracy with 5--10$\times$ fewer parameters and the lowest forgetting across both topologies.
  \item \textbf{For sequential hybrids, avoid adapting the recurrent backbone.} GDN-only or GDN-involved conditions consistently degrade target-task performance in Qwen.
  \item \textbf{For parallel hybrids, all components are viable.} SSM adaptation is safe in Falcon, though attention-only remains most parameter-efficient.
  \item \textbf{Beware destructive interference in sequential hybrids.} Combining attention and MLP without the backbone can be worse than either alone.
  \item \textbf{Prefer parallel hybrids when cross-task robustness matters.} Falcon's positive transfer contrasts sharply with Qwen's catastrophic forgetting.
\end{enumerate}

\Cref{tab:recipe} summarizes the recommended conditions from an automated selection procedure that identifies the smallest-parameter condition achieving $\geq$95\% of the full-model accuracy.

\begin{table}[t]
\centering
\caption{Practitioner recipe: recommended placement condition per model and domain. Selected as the smallest-parameter condition achieving $\geq$95\% of the full-model baseline accuracy.}
\label{tab:recipe}
\small
\begin{tabular}{@{}llcrr@{}}
\toprule
\textbf{Model} & \textbf{Domain} & \textbf{Recommended} & \textbf{Params (M)} & \textbf{$\geq$95\%?} \\
\midrule
\modelfalcon & GSM8K     & \texttt{attention\_only} & 2.21 & \checkmark \\
\modelfalcon & Code      & \texttt{attention\_only} & 2.21 & \checkmark \\
\modelfalcon & UltraChat & \texttt{attention\_only} & 2.21 & \checkmark \\
\modelqwen  & GSM8K     & \texttt{softmax\_only}   & 1.08 & \checkmark \\
\modelqwen  & Code      & \texttt{all\_layers}     & 10.82 & \checkmark \\
\modelqwen  & UltraChat & \texttt{softmax\_only}   & 1.08 & \checkmark \\
\bottomrule
\end{tabular}
\end{table}

\subsection{Relation to Concurrent Work}
\label{sec:concurrent}

Our findings complement several concurrent efforts. Galim \etal \cite{galim2025peft} showed that LoRA on linear projections outperforms LoRA on SSM-specific parameters in pure-SSM models; our results extend this to hybrid architectures and show that the attention projections specifically---not all linear projections---are the optimal target. Young's S$_0$ Tuning \cite{young2026s0tuning} takes the orthogonal approach of adapting recurrent states rather than weights; at the 4B+ scale where S$_0$ tuning excels, our component-type placement strategy could further refine which weight matrices receive LoRA alongside state tuning. The two approaches are complementary rather than competing.

\subsection{Implications for Architecture Design}
\label{sec:arch_implications}

The dramatic difference between sequential and parallel topologies in adaptation robustness has implications beyond PEFT. If parallel hybrids are fundamentally more amenable to post-training modification---exhibiting positive transfer where sequential hybrids show forgetting, and tolerating backbone adaptation where sequential hybrids are destroyed---this may favor parallel designs for models intended to be extensively fine-tuned by downstream practitioners.

\section{Limitations and Future Work}
\label{sec:limitations}

\paragraph{Scale.} Both models are sub-1B parameters. The relative importance of component types may shift at larger scales (7B+), where attention capacity is greater and the recurrent backbone may become more tolerant of perturbation. Replication at the 7--70B scale is a priority.

\paragraph{Single seed.} Experiments use a single random seed (3407). While the observed effect sizes (10--17\pp on GSM8K) substantially exceed typical LoRA variance, multi-seed replication would strengthen statistical claims. The bootstrap analysis on shared evaluation instances partially addresses this, but cannot substitute for independent training runs.

\paragraph{Fixed rank.} All experiments use $r{=}16$. The optimal placement may interact with rank: at very low ranks ($r{=}4$), the expressiveness bottleneck may shift the balance between component types. Rank ablation is a natural extension.

\paragraph{Three domains.} We evaluate on mathematics, code, and general instruction. Other domains (biomedical, legal, multilingual) may exhibit different placement sensitivities.

\paragraph{Two hybrid families.} We study one sequential (Qwen/GDN) and one parallel (Falcon/Mamba-2) hybrid. Other architectures (Jamba, Zamba, RWKV-based hybrids) may behave differently, particularly those with interleaved rather than strict sequential or parallel topologies.

\paragraph{Single PEFT method.} We study LoRA exclusively. Other PEFT methods (prefix tuning, adapters, IA$^3$) may respond differently to component-type targeting. The interaction of component placement with methods like DoRA or AdaLoRA is unexplored.

\paragraph{HumanEval at sub-1B scale.} No condition produced meaningful code generation capability (pass@1 $\leq$ 0.6\%). CodeAlpaca training may have improved code-related representations that our evaluation setup does not capture, or the sub-1B scale may simply be insufficient for functional code generation.

\section{Conclusion}
\label{sec:conclusion}

We have presented the first systematic study of component-type LoRA placement in hybrid language models. Across two architectures (sequential and parallel), three training domains, and five evaluation benchmarks, we find that:

\begin{enumerate}[nosep]
  \item The attention pathway is the optimal LoRA target in both architectures, achieving competitive or superior accuracy with 5--10$\times$ fewer trainable parameters than full-model adaptation.
  \item The recurrent backbone---the functionally dominant component during inference---is a poor adaptation target in sequential hybrids (destructive, up to $-$14.8\pp) but a viable one in parallel hybrids (constructive, up to +8.6\pp).
  \item The hybrid topology (sequential vs.\ parallel) is the primary determinant of adaptation behavior, affecting not only which components can be safely adapted but also cross-task transfer properties: parallel hybrids exhibit positive transfer where sequential hybrids show catastrophic forgetting.
\end{enumerate}

These findings establish component-type placement as a necessary consideration in PEFT configuration for hybrid architectures and provide actionable guidance: default to attention-only adaptation, and recognize that the topology of the hybrid determines whether the backbone is a viable target.

%

\section*{Declaration of competing interest}

The authors declare that they have no known competing financial interests or personal relationships that could have appeared to influence the work reported in this paper.

\section*{Data availability}

All code, configuration files, evaluation results, and trained adapters are available at \url{https://github.com/hecboar/lora-placement-hybrid}.

\section*{Acknowledgments}

This work was supported by the Valencian Research Institute for Artificial Intelligence (VRAIN) at Universitat Polit\`ecnica de Val\`encia. Computational resources were provided by Google Colab Pro and RunPod.

\bibliographystyle{elsarticle-num}

\end{document}